\documentclass{article}

\PassOptionsToPackage{numbers, compress}{natbib}
 \usepackage[dblblindworkshop, final]{neurips_2025}
\workshoptitle{Learning from Time Series for Health}

\usepackage[utf8]{inputenc} % allow utf-8 input
\usepackage[T1]{fontenc}    % use 8-bit T1 fonts
\usepackage{hyperref}       % hyperlinks
\usepackage{url}            % simple URL typesetting
\usepackage{booktabs}       % professional-quality tables
\usepackage{amsfonts}       % blackboard math symbols
\usepackage{nicefrac}       % compact symbols for 1/2, etc.
\usepackage{microtype}      % microtypography
\usepackage{xcolor}         % colors

\usepackage{graphicx}
\usepackage{algorithm}
\usepackage{algorithmic}
\usepackage{amsmath, amsthm, amssymb} %For Equations

\newtheorem*{theorem*}{Theorem}
\newtheorem*{proposition*}{Proposition}
\newtheorem{theorem}{Theorem}[section]

\newtheorem{proposition}[theorem]{Proposition}

\theoremstyle{definition}

\title{A Second-Order SpikingSSM for Wearables}

\author{%
  Kartikay Agrawal\textsuperscript{1}, Abhijeet Vikram\textsuperscript{2}, Vedant Sharma\textsuperscript{2}\and \textbf{Vaishnavi Nagabhushana\textsuperscript{1}, Ayon Borthakur\textsuperscript{1}}
  \\
  \textsuperscript{1}Sustain AI Lab, MFSDS\&AI, IIT Guwahati, Assam, India\\
  \textsuperscript{2}IISER Pune, Maharashtra, India\\
  \texttt{\{a.kartikay,n.vaishnavi,ayon.borthakur\}@iitg.ac.in} \\
  \texttt{\{vedant.sharma,abhijeet.vikram\}@students.iiserpune.ac.in} \\
}

\begin{document}

\maketitle

\begin{abstract}
Spiking neural networks have garnered increasing attention due to their energy efficiency, multiplication-free computation, and sparse event-based processing. In parallel, state space models have emerged as scalable alternatives to transformers for long-range sequence modelling by avoiding quadratic dependence on sequence length. We propose SHaRe-SSM (Spiking Harmonic Resonate-and-Fire State Space Model), a second-order spiking SSM for classification and regression on ultra-long sequences. SHaRe-SSM outperforms transformers and first-order SSMs on average while eliminating matrix multiplications, making it highly suitable for resource-constrained applications. To ensure fast computation over tens of thousands of time steps, we leverage a parallel scan formulation of the underlying dynamical system. Furthermore, we introduce a kernel-based spiking regressor, which enables the accurate modelling of dependencies in sequences of up to 50k steps. Our results demonstrate that SHaRe-SSM achieves superior long-range modelling capability with energy efficiency ($52.1\times$ less than ANN-based second-order SSM), positioning it as a strong candidate for resource-constrained devices such as wearables. 

% % 
% In addition, we conducted a systematic analysis of the impact of heterogeneity, dissipation, and conservation in resonate-and-fire SSMs.

\end{abstract}

\section{Introduction}

Spike-based deep learning has established itself as an ultra-low-power consumption and sparse computing paradigm for efficient AI in recent years. Spike-based neuromorphic hardware such as Loihi \citep{loihi, ShresthaLoihi2}, TrueNorth \citep{truenorth}, and Dynapse \citep{richter2024dynap} utilises far lower resources than conventional ANN-based designs. Apparently, most spike-based models \citep{zhou2023spikformer, 10890026, lee2025spikingtransformerspatialtemporalattention, spikingssm, stan2024learning} rely on integrate-and-fire (IF) or leaky IF neurons, which miss key biological traits like oscillations. While the biophysically detailed Hodgkin-Huxley model captures these dynamics, it is computationally prohibitive. Driven by these observations, Resonate-and-fire (RF) neurons \citep{izhikevich2001resonate}, computationally as light as IF but more expressive, have gained recent attention \citep{ShresthaLoihi2, higuchi2024balanced, c-silif}. However, RF neurons remain underexplored for very long sequence modelling. 

For sequential tasks, transformers are de facto standards \citep{attention},  but they suffer from quadratic dependence on sequence length. Alternatives, such as KV caching \citep{brandon2024reducing} and memory updates \citep{behrouz2024titans}, reduce overhead but lack the simplicity of RNNs. State space models (SSMs) \citep{s4, s5, mamba} and their spiking variants \citep{stan2024learning, spikingssm, pspikessm} bridge this gap. Yet current spiking SSMs struggle with very long sequences.

LinOSS \citep{rusch2025oscillatory,dlinoss}, a second-order SSM, addresses this by using stable discretisations with diagonal state matrices, achieving state-of-the-art results on long-range tasks. However, it lacks spike-based communication, which is crucial for energy efficiency \citep{ShresthaLoihi2, Imam2020-jm}, such as demanded by a battery-driven wearable. Hence, in this work, we introduce SHaRe-SSM, a second-order spiking SSM designed for extremely long-range tasks and energy-efficient edge AI. Our contributions are: 
(1) A fully spike-based second-order SSM, without ANN nonlinearities (GeLU, GLU, or GSU).
(2) A compatible parallel scan algorithm for fast training and inference.
Superior accuracy over first-order SSMs on long-sequence classification, with higher efficiency than ANN-based second-order SSMs.
(3) An extension to regression via a convolving kernel, outperforming first-order SSMs on 50K-length tasks. 
% A study on discretisation methods and network heterogeneity in SHaRe-SSM.
\section{Methods}

\subsection{Network Description}

The Resonate-and-Fire (RF) neuron \citep{izhikevich2001resonate} provides a closer approximation to the Hodgkin-Huxley (HH) model than the Leaky Integrate-and-Fire (LIF) neuron by capturing subthreshold resonance through a 2D linear system with complex eigenvalues. This enables oscillatory dynamics and frequency selectivity, traits absent in LIF but observed in HH neurons. To simplify implementation, second-order real-valued variants such as the Harmonic RF (HRF) \citep{hrf,higuchi2024balanced} reformulate RF as a real harmonic oscillator while preserving oscillatory behaviour without requiring complex initialisation. RF’s complex dynamics make it more biophysically realistic, and recent work \citep{higuchi2024balanced} shows that RF-based models perform best without membrane resets. Building on these insights, we integrate HRF into {LinOSS's} SSM framework \cite{rusch2025oscillatory}, compute dynamics linearly via parallel scans, and use a spike function as the activation to propagate spikes.
% \begin{align}
% \label{eq:hrf}
% \begin{split}
% u'(t) &= - \omega^2 v(t) -2b u(t)  + x(t) \\
% v'(t) &= u(t) \\
% \end{split}
% \end{align}
% where $ v(t)$ denote the hidden state, $u(t)$ as the time derivative of $v(t)$, $x(t)$ is the input spikes to the neuron, \( \omega \in \mathbb{R} \) is the frequency parameter and \( b \in \mathbb{R} \) is the damping coefficient. Spiking Function: $\Theta \in \{0,1\}$ i.e. spikes are emitted if $u \ge \theta$.
% \begin{figure*}[ht]
% \centering
% \includegraphics{final_SHaRe-SSM.png} 
% \caption{Description of SHaRe-SSM model. SHaRe-SSM consists of a SHaRe-SSM neuron representing multiple states (Eq. \ref{alg: SHaRe}) followed by a Linear layer repeated for N blocks. Spikes are encoded with a Spike Encoder and decoded using a Spike Decoder.}
% \label{fig: model}
% \end{figure*}
This original HRF formulation is limited to a single neuron. We propose formulating HRF in an SSM framework (SHaRe-SSM). This approach helps in capturing better dynamics. The input spikes are multiplied by a weight matrix, and we propagate spikes from the hidden state $v(t)$. Moreover, removing the damping parameter in such second-order approximations enhances the neuron's ability to capture long-range temporal dependencies by preserving longer oscillations   \cite{rusch2025oscillatory}. We define the SHaRe-SSM model by: 
\begin{align} \label{eq: SHaRe-SSM}
\begin{split}
u'(t) &= -\Omega v(t) + B x (t) \\
v'(t) &= u(t) \\
z(t) &=\Theta(v(t)-\theta_C)\\
\end{split}
\end{align}
where \( u(t), v(t) \in \mathbb{R}^p \) denote the hidden states, \( y(t) \in \mathbb R^h \) the output, and \( x(t) \in \mathbb{R}^h \) the input spike signal. The system is defined by weights \( \Omega \in \mathbb{R}^{p\times p} \) which is diagonal, \( B \in \mathbb{R}^{p \times h} \), \( C \in \mathbb{R}^{h \times p} \), \( D \in \mathbb{R}^{h} \), and an output learnable spiking parameter \( \theta_C \in \mathbb{R}^p \) such that spikes are emitted if $v \geq \theta_C$. These thresholds ($\theta$) are learned using a step-double Gaussian surrogate gradient \citep{neftci2019surrogate,higuchi2024balanced}. Since \citep{higuchi2024balanced} demonstrated that HRF performs well without a reset, we adopted the same strategy here. This eliminates the need for sequential processing, allowing the model to be implemented across time as an activation function. Each SHaRe-SSM block is comprised of a SHaRe-SSM neuron, followed by a linear layer, and a spike function (IF neuron with no reset).

\subsection{Encoder and Decoder}
For SHaRe-SSM, we design an encoder for encoding the input signal into spike trains (see Algorithm \ref{alg: SHaRe}). Herein, the inputs every timestep are passed through a learnable linear layer. Finally, an IF neuron with no reset, i.e., a spike function, generates spikes into the SHaRe-SSM block. Notably, this is a data-dependent trainable encoder (without requiring us to specify whether to utilise rate coding or any other encoding variants). Moreover, it utilises neuronal heterogeneity and is instantaneous. The decoder projects the spikes back using a linear Layer with the output dimension equal to the number of classes. For regression, we also propose convolving the output with a learnable filter.

% For learning via backpropagation, we used a surrogate gradient descent \cite{neftci2019surrogate, higuchi2024balanced}.
% (details in Regression with Spikes \ref{sec: Regression}). 

\subsection{Discretisation methods}

Euler Forward (explicit) discretisation causes divergence over time, making the model unstable. In contrast, IMEX discretisation remains stable and preserves energy, while the IM scheme, though dissipative, is also stable. Below, we analyse the generalizability of IM and IMEX for long sequences. IMEX is particularly promising for regression: as shown in \citet{rusch2025oscillatory}, removing damping yields a Hamiltonian system \citep{arnold1989mathematical}, where energy conservation is guaranteed under symplectic discretisations. We also evaluate the non-conservative but empirically stable IM scheme. Our formulation adopts a second-order system with position-like state \( u_n \) and velocity-like state \( v_n \).
The general form of discrete updates is denoted by:
\begin{align}
u_n &= u_{n-1} + \Delta t (-\Omega v_{\star} + B x_n), \\
v_n &= v_{n-1} + \Delta t\, u_n, \\
s_n &= M s_{n-1} + F_n,
\end{align}
% \begin{align}
% \label{eq:general_model}
% \end{align}
where the choice of \( v_\star \) distinguishes the discretisation schemes: \( v_\star = v_n \) for IM (implicit), and \( v_\star = v_{n-1} \) for IMEX (implicit-explicit).
% For the IM scheme, we define \cite{rusch2025oscillatory}:
% \begin{align*}
% M^{\text{IM}} &= 
% \begin{pmatrix}
%     S & -S \Delta t \Omega \\
%     S \Delta t & S
% \end{pmatrix}, \quad
% F_n^{\text{IM}} = 
% \begin{pmatrix}
%     S \Delta t B x_n \\
%     S \Delta t^2 B x_n
% \end{pmatrix},
% \end{align*}
% with \( S = (I + \Delta t^2 \Omega)^{-1} \), acting as a Schur complement that suppresses high-frequency components and ensures eigenvalues of \( M^{\text{IM}} \) remain bounded within the unit circle.

% And, for the IMEX scheme, we use \cite{rusch2025oscillatory}:
% \begin{align*}
% M^{\text{IMEX}} &= 
% \begin{pmatrix}
%     I & -\Delta t \Omega \\
%     \Delta t I & I - \Delta t^2 \Omega
% \end{pmatrix}, \quad
% F_n^{\text{IMEX}} = 
% \begin{pmatrix}
%     \Delta t B x_n \\
%     \Delta t^2 B x_n
% \end{pmatrix}.
% \end{align*}
Unlike IM, the eigenvalues of \( M^{\text{IMEX}} \) lie near the unit circle, preserving oscillatory energy over long horizons. This makes IMEX especially suitable for modelling long sequences, where maintaining temporal structure is crucial.

\begin{algorithm}
\caption{SHaRe-SSM Algorithm}\label{alg: SHaRe}
\begin{algorithmic}
\REQUIRE Input sequence $x$
\ENSURE $N$-blocks, spike function $\Theta$ , output sequence $o$
% \STATE $x^0 \gets \Theta(W_{enc}x+b_{enc})$ \COMMENT{Encode input sequence into spikes}
\STATE $x^0 \gets \textit{Encoder}(x)$ \COMMENT{Encode input sequence into spikes}
\FOR{$n=1,\dots,N$}
\STATE $z^n \gets$ solution of HRF in \eqref{eq: SHaRe-SSM} with input $x^{n-1}$ via parallel scan aggregated
% \STATE ${z^n \gets \Theta( z^n -\theta^n_C)}$ 
\STATE $y^n \gets Cz^n+Dx^{n-1}$ \COMMENT{Weighted spike mixing in \eqref{eq: SHaRe-SSM}}
\STATE $y^n \gets \Theta(y^n-\theta_D^n)$
% \STATE $y^n \gets drop(\Theta(y^n-\theta_D^n))$
\STATE $y^n \gets Linear(y^n)$ 
% \STATE $y^n \gets \textit{BN}(y^n))$
\STATE $y^n \gets \Theta(y^n-\theta^n)$
\STATE $x^{n} \gets y^n + x^{n-1}$ \COMMENT{Spike mixing}
\ENDFOR
% \STATE $o \gets W_{dec}\sum_{i=0}^Ly^i+b_{dec}$ %\COMMENT{Decode final LinOSS block output}
% \STATE {$o \gets o \ast k$} \COMMENT{convolution with LI filter}
\STATE $o \gets \textit{Decoder}(x^N)$ \COMMENT{Decode spikes}
\end{algorithmic}
\end{algorithm}

\section{Emperical Results}
\subsection{Human Activity Recognition}
Our SNN model is capable of learning multiple states without any sequential recurrence, as it does not have a reset mechanism. Such a design enables parallelisation and reduced energy consumption. Hence, it is well-suited for wearable devices. Although our model is designed for long-range sequence datasets, we also evaluated its performance on Human Activity Recognition(HAR) datasets: UCI-HAR \citep{anguita2013public} and SHAR datasets \citep{micucci2017unimib}. For 30 subjects, UCI-HAR comprises 10.3k instances of 6 activities (walking, walking upstairs/downstairs, sitting, standing, lying) captured with a 3-axis accelerometer and gyroscope (50 Hz) on a Samsung Galaxy SII, while UniMB SHAR includes 11.7k instances of 17 activities (9 daily, 6 fall) recorded with a 3-axis accelerometer ($\leq 50$ Hz) on a Samsung Galaxy Nexus I9250. SHaRe-SSM achieves an accuracy of 99\% and outperforms the best-performing model ~\citep{li2022wearablebased} for UCI-HAR by 0.2\%. For SHAR, our model achieves 92.7\%, which is better than non-spiking models in terms of energy and performance, and is competitive with spiking models, as it outperforms SpikeDCL by 0.6\% and falls behind SpikeDCL by 1.2\% (Refer to Table \ref{tab:har_comparison}).
% and HHAR~\cite{stisen2015smart}
% \begin{table}[ht]
% \centering
% \caption{Performance comparison between different networks on HAR datasets: UCI-HAR\cite{anguita2013public}, and SHAR~\cite{micucci2017unimib},  (DCL:DeepConvLSTM, Trans:Transformer)}
% \label{tab:har_comparison}
% \resizebox{\linewidth}{!}{%
% \begin{tabular}{lcccccccc}
% \hline
% \textbf{Model} & {CNN} & {DCL} & {LSTM} & {Trans} & {SpikeCNN} & {SpikeDCL} & {Our(IM)} & {Our(IMEX)}\\ 
% \hline
% \textbf{SNN} & {N} & {N} & {N} & {N} & {Y} & {Y} & {Y} & {Y}\\ 
% \hline
% UCIHAR& 96.3$\pm$0.1 & 97.9$\pm$0.3 & 82.4$\pm$4.0 & 96.0$\pm$0.3 & 96.4$\pm$0.2 & \textbf{98.9$\pm$0.3} & \textbf{99.02$\pm$0.3}
% &\textbf{96.93$\pm$0.5}
% \\
% SHAR& 92.4$\pm$0.5 & 90.8$\pm$1.0 & 83.9$\pm$0.9 & 83.2$\pm$0.7 & \textbf{94.0$\pm$0.3} & 92.1$\pm$0.8 & \textbf{92.76$\pm$0.07}& \textbf{89.08$\pm$0.7} \\
% % HHAR& 96.2$\pm$0.1 & 97.2$\pm$0.2 & 95.6$\pm$0.2 & 95.8$\pm$0.2 & 96.2$\pm$0.1 & \textbf{97.5$\pm$0.1} & &\\
% \hline
% \end{tabular}
% }
% \end{table}
\begin{table}[ht]
\centering
% ,  (DCL: DeepConvLSTM, Trans: Transformer)}
\begin{tabular}{lccc}
\hline
\textbf{Model} & \textbf{SNN} & \textbf{UCIHAR} & \textbf{SHAR} \\ 
\hline
CNN & No & 96.3$\pm$0.1 & 92.4$\pm$0.5 \\ 
DeepConvLSTM & No &97.9$\pm$0.3 & 90.8$\pm$1.0 \\ 
LSTM & No & 82.4$\pm$4.0 & 83.9$\pm$0.9 \\ 
Transformers & No & 96.0$\pm$0.3 & 83.2$\pm$0.7 \\ 
SpikeCNN & \textbf{Yes} & 96.4$\pm$0.2 & \textbf{94.0$\pm$0.3} \\ 
SpikeDCL & \textbf{Yes} & \underline{98.9$\pm$0.3} & 92.1$\pm$0.8 \\ 
Ours (IM) & \textbf{Yes} &\textbf{99.02$\pm$0.3} & \underline{92.8$\pm$0.1} \\ 
Ours (IMEX) & \textbf{Yes} &{96.9$\pm$0.5} & 89.1$\pm$0.7 \\ 
\hline
\end{tabular}
\caption{Performance comparison (Accuracy \%) between different networks on Human Activity Recognition datasets: UCI-HAR\citep{anguita2013public}, and SHAR~\citep{micucci2017unimib}}
\label{tab:har_comparison}
\end{table}

\subsection{Very-Long Range Interactions}

% \label{sec: very long range interactions}
Wearable devices operate over long durations, demanding robust prediction on extended sequences. Spiking SSMs, with their high throughput and energy efficiency, are well-suited for this setting. To evaluate performance on very long temporal dependencies, we study the longest benchmark datasets: the Eigenworms dataset (17,984 sequences) for classification and the PPG-DaLiA dataset (49,920 sequences) for regression. Following the hyperparameter protocol of \citet{rusch2025oscillatory} and employing Bayesian search \citep{akiba2019optuna}, we ensure an optimised training. Eigenworms, a subset of the UEA archive, represent C. elegans motion through six eigenworm features to distinguish wild-type behaviour from that of other mutants. PPG-DaLiA involves heart rate prediction from wrist-worn sensor data collected from 15 individuals over 150 minutes at 128 Hz.

For the Long-range classification task on EigenWorms, we observe that our model performs comparably to LinOSS with just $2.2\%$ drop in performance for IM discretisation, and significantly outperforms the IMEX counterpart by $10\%$. We observe an additional 82.6\% energy improvement for the IMEX discretisation (Fig. \ref{fig: energy computation}). Our model outperforms all previous models, especially first-order SSMs such as LRU and S6 by $5\%$ and $7.8\%$ respectively.

We present the first regression results of a spiking SSM on extremely long sequences (up to 50k). To address the limited output range of spiking neurons, we introduce a kernel-based spiking regressor with a learnable temporal kernel. As shown in Table \ref{tab: ew_ppg}, SHaRe-SSM models consistently outperform all first-order SSMs. SHaRe-SSM-IMEX surpassing Mamba by $0.016$ MSE, demonstrating both the energy-efficient nature of IMEX discretisation and the strong representation power of resonating neurons. Despite their efficiency, our models are only at most $0.027$ MSE below second-order SSMs.

We observed from \citet{rusch2025oscillatory} that IM discretisation outperforms IMEX for all datasets for classification and vice versa for regression. This trend is observed similarly for our model.  Additionally, \citet{hu2024state} demonstrated that linear SSM-based models outperform RNN-based models for chaotic systems. \citet{pourcel2025learning} observed similar trends for the EigenWorms dataset, where the non-linear dynamics from \citet{unicornn} suffer a tragic drop in performance; however, for the PPG dataset, it outperforms its linear counterpart, and only lags behind LinOSS-based models.
% The dataset is split by a sliding window of step size 4992 and then divided into a 70/15/15 split for each individual. 

% \begin{table}[!ht]
% \caption{Mean and Standard Deviation reported for longest sequences: Accuracy for EigenWorms, and Mean-Squared Error (MSE $\times 10^{-2}$) for PPG-DaLiA dataset across five training runs for the best model. (LNCDE: Log-NCDE, OSS: LinOSS)}
% \label{tab: ppg}
% \resizebox{\linewidth}{!}{
% \centering
% \begin{tabular}{|c|c|cccccccc|c|}
% \hline
% Data & Metric & NRDE & NCDE & LNCDE & LRU & S5 & S6 & Mamba & OSS & Ours\\
% \hline
% & SNN & N & N & N & N & N & N & N & N & Y \\
% \hline
% EW & ACC ($\uparrow$)& 83.9 & 75.0 & 85.6 & 87.8 & 81.1 & 85.0 & 70.9 & \bf{95.0} & \bf{92.8} \\(18k) & (Classification) & $\pm$7.3 & $\pm$3.9 & $\pm$5.1 & $\pm$2.8 & $\pm$3.7 & $\pm$16.1 & $\pm$15.8 & \bf{$\pm$4.4} & \bf{$\pm$3.3} \\
% \hline
% PPG & MSE ($\downarrow$)& 9.9 & 13.5 & 9.6 & 12.2 & 12.6 & 12.9 & 10.7 & \bf{6.4} & \bf{9.1} \\
% (50k) & (Regression)& $\pm$1.0 & $\pm$0.7 & $\pm$0.6 & $\pm$0.5 & $\pm$1.3 & $\pm$2.1 & $\pm$2.2 & \bf{$\pm$0.2} & \bf{$\pm$0.2} \\
% \hline
% \end{tabular}}
% \end{table}

\begin{table}[h]
\centering
\begin{tabular}{lccccc}
\hline
Method & Integrator & SNN & EW (ACC $\uparrow$) & PPG (MSE $\downarrow$) \\
\hline
NRDE\cite{nrde}   && No & 83.9 $\pm$ 7.3 & 9.9 $\pm$ 1.0 \\
NCDE\cite{ncde}   && No & 75.0 $\pm$ 3.9 & 13.5 $\pm$ 0.7 \\
Log-NCDE\cite{walker2024log}  && No & 85.6 $\pm$ 5.1 & 9.6 $\pm$ 0.6 \\
LRU\cite{lru}    && No & 87.8 $\pm$ 2.8 & 12.2 $\pm$ 0.5 \\
S5\cite{s5}     && No & 81.1 $\pm$ 3.7 & 12.6 $\pm$ 1.3 \\
S6\cite{mamba}     && No & 85.0 $\pm$ 16.1 & 12.9 $\pm$ 2.1 \\
Mamba\cite{mamba}  && No & 70.9 $\pm$ 15.8 & 10.7 $\pm$ 2.2 \\
\hline
LinOSS\cite{rusch2025oscillatory}  & IM & No & \color{red}\textbf{95.0 $\pm$ 4.4} & \color{blue}\textbf{7.5 $\pm$ 0.5} \\
\textbf{Ours}    & IM &  \textbf{Yes} & \color{blue}\textbf{92.8 $\pm$ 3.3} &  {$11.8 \pm  0.9$}\\
\hline
RHEL-Lin\cite{pourcel2025learning}    &IMEX &   No & 75.0 $\pm$ 9.9   & 9.5$\pm$ 1.0 \\
RHEL-Nonlin\cite{pourcel2025learning} & IMEX &   No & 50.1 $\pm$ 6.7   & \color{violet}\textbf{8.4$\pm$ 0.5} \\
LinOSS\cite{rusch2025oscillatory}  & IMEX &   No & {80.0 $\pm$ 4.4} & \color{red}\textbf{6.4 $\pm$ 0.2} \\
\textbf{Ours}    & IMEX &  \textbf{Yes} & \color{violet}\textbf{90.0 $\pm$ 5.7} & {9.1 $\pm$ 0.2} \\
\hline
\end{tabular}
\caption{Mean and Standard Deviation reported for longest sequences: Accuracy for EigenWorms, and Mean-Squared Error (MSE $\times 10^{-2}$) for PPG-DaLiA dataset across five training runs for the best model (Top three models are highlighted in  {\color{red}\textbf{red}},  {\color{blue}\textbf{blue}},  {\color{violet}\textbf{violet}} respectively).}
\label{tab: ew_ppg}
\end{table}

\subsubsection{Energy computation}

\begin{figure}[!ht]
\centering
\includegraphics{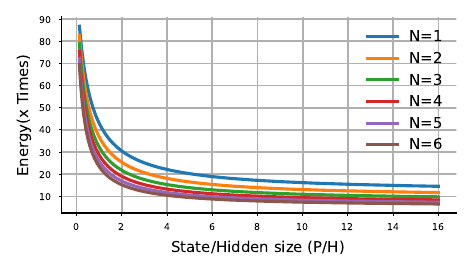} % Reduce the figure size to slightly narrower than the column.
\caption{Variation of energy computation of a LinOSS Block \citep{rusch2025oscillatory} to the SHaRe-SSM Block for EigenWorms with respect to State (P), Hidden size (H).}
\label{fig: energy computation}
\end{figure}

% For a fair comparison, let's keep hyperparameters fixed to input dim(I) = $6$, hidden dim(H) = $16$, output dim(O) = $5$, state dim(P) = $64$, num blocks(N) = $6$, and L denotes sequence length
% % include time=$true$, lr = $0.001$.

In this section, we assess and compare the energy computation for LinOSS and SHaRe-SSM for similar hidden-size(H), state-size(P), and num-blocks(N) on the EigenWorms dataset. From Figure \ref{fig: energy computation}, we observe that for different state-to-hidden ratios, we get different energy numbers. For the EigenWorms dataset, the average spike rate per time step per block is $0.42$. 
% For a lower P/H ratio, we yield high energy efficiencies ($\sim 82.6\times$). 
% Even for a high ratio, our model block is $\sim6.6\times$ more energy-efficient than a LinOSS Block for EigenWorms. 
% Our best performing model (as per Table \ref{tab: ew_ppg}) is $\textbf{52.1}\times$ more energy efficient than an equivalent LinOSS model.
We observe our best performing IMEX model to be  $\bf {\sim 52.1\times}$ energy efficient than an equivalent LinOSS model. Furthermore, since our model also outperforms LinOSS by 10\%, it can captures state dynamics better than LinOSS.

\section{Discussion}
We propose SHaRe-SSM, an energy-efficient second-order spiking SSM built with harmonic resonate-and-fire neurons, along with a learnable encoder, decoder, and parallel scan method. Unlike prior similar SSMs, SHaRe-SSM is fully spike-based without GeLU/GLU and is tailored for very long-sequence modelling. The model performs well on HAR datasets while consuming significantly less energy, and also achieves superior classification and regression performance on 18k EigenWorms and 50k-length sequences PPG datasets, respectively. Hence, it is ideally suited for use in wearable devices in healthcare. Future work will focus on the deployment of SHaRe-SSM for edge AI using Intel Loihi2 \citep{ShresthaLoihi2}.
\bibliographystyle{plainnat}
\bibliography{neurips_2025}

\appendix
\section{Technical Appendix}
\subsection{Background Theory}
To study the theoretical properties of a second-order ODE in SHaRe-SSM, we take inspiration from \citet{rusch2025oscillatory}. Our model, like theirs, can be formulated as energy-conserving and possessing dissipative attributes.

\subsubsection{Implicit discretisation (IM):}  
We consider the implicit (backward Euler) discretisation of a second-order system involving a position-like state $u_n$ and a velocity-like state $v_n$, similar to \citet{rusch2025oscillatory}. The implicit scheme is known to introduce additional dissipative terms, which contribute to the stability of the dynamics, particularly in the presence of stiffness.

The discretised updates are given by:
\begin{align*}
u_n &= u_{n-1} + \Delta t \left(- \Omega v_n + B x_n \right), \\
v_n &= v_{n-1} + \Delta t\, u_n,
\end{align*}
where $\Omega$ is a diagonal matrix of oscillation frequencies and $B$ is an input projection matrix. Note that both $u_n$ and $v_n$ are evaluated at the future timestep, in contrast to explicit methods.

Letting the concatenated state be $s_n$, the above system can be written compactly as:
\[
M s_n = s_{n-1} + F_n,
\]
where
\begin{align*}
M &= \begin{pmatrix}
    I & \Delta t \Omega \\
    -\Delta t I  & I 
    \end{pmatrix}, \qquad
F_n = \begin{pmatrix}
     \Delta t Bx_n \\ 0   
    \end{pmatrix}.
\end{align*}

To obtain an explicit update rule, we algebraically solve the coupled system by introducing the matrix inverse $S = (I + \Delta t^2 \Omega)^{-1}$. Substituting and simplifying yields:
\begin{align}
\label{eq:im_model}
s_n = M^{\text{IM}} s_{n-1} + F_n^{\text{IM}},
\end{align}
where
\begin{align*}    
M^{\text{IM}} &= \begin{pmatrix}
    S & -S \Delta t \Omega \\
    S \Delta t & S
    \end{pmatrix}, \qquad
F_n^{\text{IM}} = \begin{pmatrix}
     S \Delta t B x_n \\ S \Delta t^2 B x_n
    \end{pmatrix}.
\end{align*}

This formulation highlights the stabilising effect of the implicit method: the matrix $S = (I + \Delta t^2 \Omega)^{-1}$ is a Schur complement that acts as a preconditioner, suppressing high-frequency components. Consequently, the eigenvalues of $M^{\text{IM}}$ remain bounded within the unit circle for a wide range of $\Delta t$, leading to improved numerical stability. The Schur complement can be computed in $\mathcal{O}(m)$ instead of the typical $\mathcal{O}(m^3)$ operations using Gauss-Jordan elimination.

{Proposition A.1 is rephrased from proposition 3.1 of Rusch and Rus. \citep{rusch2025oscillatory}}
\begin{proposition}
\label{prop:ev_im}
Let $M^{\text{IM}} \in \mathbb R^{2p\times 2p}$ be the hidden-to-hidden weight matrix of the implicit model SHaRe-SSM-IM \eqref{eq:im_model}. We assume that $\Omega_{j} \geq 0$ for all diagonal elements $j=1,\dots,p$ of $\Omega$, and further that $\Delta t>0$. Then, the complex eigenvalues of $M^{\text{IM}}$ are given as, 
\begin{equation*}
    \lambda_{j_{1,2}} = \frac{1}{1+\Delta t^2\Omega_{j}} \pm \Delta t \frac{\sqrt{\Omega_{j}}}{1+\Delta t^2\Omega_{j}}
\end{equation*}
with \( \lambda_{j_1} = \overline{\lambda_{j_2}} \).
Moreover, the spectral radius $\rho(M^{\text{IM}})$ is bounded by $1$, i.e., $|\lambda_j| 
\leq 1$ for all $j=1,\dots,p$.
\end{proposition}
\begin{proof}
The matrix \( M^{\text{IM}} \in \mathbb{R}^{2p \times 2p} \) is defined as
\[
M^{\text{IM}} =
\begin{bmatrix}
S & -\Delta t\, \Omega S \\
\Delta t\, S & S
\end{bmatrix},
\]
where \( S = (I + \Delta t^2 \Omega)^{-1} \), and \( \Omega \in \mathbb{R}^{p \times p} \) is diagonal with non-negative entries \( \Omega_j \geq 0 \) for all \( j = 1, \dots, p \).

To determine the eigenvalues, we compute the characteristic polynomial:
\[
\det(M^{\text{IM}} - \lambda I) =
\begin{vmatrix}
S - \lambda I & -\Delta t\, \Omega S \\
\Delta t\, S & S - \lambda I
\end{vmatrix}.
\]
Using block Gaussian elimination, we subtract \( (\Delta t\, S)(S - \lambda I)^{-1}(-\Delta t\, \Omega S) \) from the lower-right block, giving
\[
= \det(S - \lambda I) \cdot \det\left(S - \lambda I + \Delta t^2\, \Omega S^2 (S - \lambda I)^{-1} \right).
\]

Since \( S \) and \( \Omega \) are diagonal and commute, this expression decouples elementwise. Let \( s_j = \frac{1}{1 + \Delta t^2 \Omega_j} \) be the \( j \)-th diagonal element of \( S \). Then for each \( j = 1,\dots,p \), the scalar characteristic equation becomes
\[
(s_j - \lambda)^2 + \Delta t^2 \Omega_j s_j^2 = 0.
\]

Solving this quadratic gives the eigenvalue pair
\[
\lambda_{j_{1,2}} = s_j \pm i\, \Delta t\, s_j \sqrt{\Omega_j},
\]
where \( \lambda_{j_1} = \overline{\lambda_{j_2}} \), i.e., the pair are complex conjugates.

To compute their magnitude:
\[
|\lambda_{j_{1,2}}|^2 = s_j^2 (1 + \Delta t^2 \Omega_j) = \frac{1}{1 + \Delta t^2 \Omega_j} \leq 1.
\]

Hence, all eigenvalues lie on or inside the unit circle in the complex plane, and the spectral radius satisfies \( \rho(M^{\text{IM}}) \leq 1 \), as claimed.
\end{proof}

{Proposition A.2 is adapted from proposition 3.2 of Rusch and Rus. \citep{rusch2025oscillatory}}
\begin{proposition}
\label{prop:ev_im_expected}
Let \( \{\lambda_j\}_{j=1}^{2p} \) denote the eigenvalues of the hidden-to-hidden matrix \( M^{\text{IM}} \in \mathbb{R}^{2p \times 2p} \) of the SHaRe-SSM-IM model~\eqref{eq:im_model}. Suppose the diagonal entries of \( \Omega \in \mathbb{R}^{p \times p} \) are independently drawn as \( \Omega_j \sim \mathcal{U}([0, \Omega_{\max}]) \), with \( \Omega_{\max} > 0 \). Then, the \( N \)-th moment of the magnitude of the eigenvalues is given by
\[
\mathbb{E}(|\lambda_j|^N) = \frac{(1 + \Delta t^2 \Omega_{\max})^{1 - \frac{N}{2}} - 1}{\Delta t^2 \Omega_{\max} (1 - \frac{N}{2})}, \quad \forall j = 1, \dots, 2p.
\]
\end{proposition}

\begin{proof}
From Proposition~\ref{prop:ev_im}, each eigenvalue of \( M^{\text{IM}} \) has magnitude
\[
|\lambda_j| = \sqrt{s_j} = \sqrt{ \frac{1}{1 + \Delta t^2 \Omega_j} }.
\]
The \( N \)-th moment of the magnitude is thus.
\[
\mathbb{E}(|\lambda_j|^N) = \mathbb{E} \left[ \left( \frac{1}{1 + \Delta t^2 \Omega_j} \right)^{\frac{N}{2}} \right].
\]

Applying the law of the unconscious statistician and the uniform distribution of \( \Omega_j \sim \mathcal{U}([0, \Omega_{\max}]) \), we write:
\[
\mathbb{E}(|\lambda_j|^N) = \frac{1}{\Omega_{\max}} \int_0^{\Omega_{\max}} (1 + \Delta t^2 x)^{-\frac{N}{2}} \, dx.
\]

Substituting \( u = 1 + \Delta t^2 x \), so that \( du = \Delta t^2 \, dx \), the limits change from \( x = 0 \) to \( x = \Omega_{\max} \), corresponding to \( u = 1 \) to \( u = 1 + \Delta t^2 \Omega_{\max} \). The integral becomes:
\[
\mathbb{E}(|\lambda_j|^N) = \frac{1}{\Delta t^2 \Omega_{\max}} \int_1^{1 + \Delta t^2 \Omega_{\max}} u^{-\frac{N}{2}} \, du.
\]

This evaluates to
\[
\mathbb{E}(|\lambda_j|^N) = \frac{(1 + \Delta t^2 \Omega_{\max})^{1 - \frac{N}{2}} - 1}{\Delta t^2 \Omega_{\max}(1 - \frac{N}{2})},
\]
\end{proof}

We can observe from proposition \ref{prop:ev_im_expected} that even though the spectral radius of eigenvalues is smaller than one (proposition \ref{prop:ev_im}), it is large enough to capture long-range dependencies even for very long-range sequences, even for $\Omega_{\max}=1,\Delta t=1$. Hence, we initialize \( \Omega_j \sim \mathcal{U}([0, 1]),\Delta t_j \sim \mathcal{U}([0, 1])\)

\subsubsection{Implicit-Explicit discretisation (IMEX):}  
We also utilise an implicit-explicit (IMEX) scheme for discretising the second-order harmonic oscillator system, similar to \citet{rusch2025oscillatory}. IMEX methods treat the stiff terms implicitly and the non-stiff or input terms explicitly, resulting in a balanced scheme that enables stable yet undamped oscillations. As shown in \citet{rusch2025oscillatory}, such schemes preserve the total energy of the system and therefore are particularly well-suited for learning long-range sequential patterns without introducing artificial dissipation.

The update equations under the IMEX discretisation are given by:
\begin{align*}
    u_n &= u_{n-1} + \Delta t \left(  - \Omega v_{n-1} + B x_n \right), \\
    v_n &= v_{n-1} + \Delta t\, u_n,
\end{align*}
where the velocity update depends implicitly on the newly computed $u_n$, while the force term $-\Omega v_{n-1} + B x_n$ is evaluated using previous state values.

Defining the state vector as $s_n$, we can rewrite the update in matrix form:
\begin{align*}
M s_n = M_1 s_{n-1} + F_n,
\end{align*}
where the matrices $M$, $M_1$, and input vector $F_n$ are:
\begin{align*}
M &= \begin{pmatrix}
    I & 0 \\
    -\Delta t I & I 
    \end{pmatrix},
M_1 = \begin{pmatrix}
    I & -\Delta t \Omega \\
    0 & I 
    \end{pmatrix}, 
F_n = \begin{pmatrix}
     \Delta t B x_n \\ 0   
    \end{pmatrix}.
\end{align*}

Multiplying both sides by $M^{-1}$ yields the closed-form update:
\begin{align}
\label{eq:imex_model}
s_n = M^{\text{IMEX}} s_{n-1} + F_n^{\text{IMEX}},
\end{align}
where the transition matrix and input vector are given by:
\begin{align*}
M^{\text{IMEX}} &= \begin{pmatrix}
    I & -\Delta t \Omega \\
    \Delta t I & I - \Delta t^2 \Omega
    \end{pmatrix}, \qquad
F_n^{\text{IMEX}} = \begin{pmatrix}
     \Delta t B x_n \\ \Delta t^2 B x_n
    \end{pmatrix}.
\end{align*}

{Proposition A.3 is adapted from proposition E.1 of Rusch and Rus. \citep{rusch2025oscillatory}}

\begin{proposition}
\label{prop:ev_imex}
Let \( M^{\text{IMEX}} \in \mathbb{R}^{2p \times 2p} \) be the hidden-to-hidden weight matrix of the implicit-explicit model SHaRe-SSM-IMEX~\eqref{eq:imex_model}. Suppose that \( \Omega \in \mathbb{R}^{p \times p} \) is diagonal with strictly positive entries \( \Omega_j > 0 \) for all \( j = 1, \dots, p \), and that the time step satisfies \( 0 < \Delta t \leq \max_j \left( \frac{2}{\sqrt{\Omega_j}} \right) \). Then, the eigenvalues of \( M^{\text{IMEX}} \) are given by
\[
\lambda_{j_{1,2}} = \frac{1}{2} (2 - \Delta t^2 \Omega_j) \pm \frac{1}{2} \sqrt{ \Delta t^2 \Omega_j (4 - \Delta t^2 \Omega_j)},
\]
with \( \lambda_{j_1} = \overline{\lambda_{j_2}} \). Moreover, the eigenvalues lie on the complex unit circle, i.e., \( |\lambda_j| = 1,\quad \forall  j = 1, \dots, p \).
\end{proposition}

\begin{proof}
The matrix \( M^{\text{IMEX}} \in \mathbb{R}^{2p \times 2p} \) has the block form
\[
M^{\text{IMEX}} =
\begin{bmatrix}
I & -\Delta t\, \Omega \\
\Delta t\, I & I
\end{bmatrix}
\begin{bmatrix}
(I + \Delta t^2 \Omega)^{-1} & 0 \\
0 & (I + \Delta t^2 \Omega)^{-1}
\end{bmatrix},
\]
So the effective system matrix becomes
\[
M^{\text{IMEX}} =
\begin{bmatrix}
S & -\Delta t\, \Omega S \\
\Delta t\, S & S
\end{bmatrix}, \quad \text{where } S = (I + \Delta t^2 \Omega)^{-1}.
\]

We analyse the characteristic polynomial:
\[
\det(M^{\text{IMEX}} - \lambda I) =
\begin{vmatrix}
S - \lambda I & -\Delta t\, \Omega S \\
\Delta t\, S & S - \lambda I
\end{vmatrix}.
\]
Using block elimination, we simplify:
\[
= \det(S - \lambda I)^2 + \Delta t^2 \Omega S^2.
\]
Since all matrices are diagonal, the problem decouples elementwise. Let \( s_j = \frac{1}{1 + \Delta t^2 \Omega_j} \) for each \( j = 1, \dots, p \). Then, for each \( j \), the characteristic polynomial becomes
\[
\lambda^2 - (2s_j)\lambda + (s_j^2 + \Delta t^2 \Omega_j s_j^2) = \lambda^2 - (2 - \Delta t^2 \Omega_j)\lambda + 1 = 0.
\]

Solving this gives the eigenvalue pair
\[
\lambda_{j_{1,2}} = \frac{1}{2} (2 - \Delta t^2 \Omega_j) \pm \frac{1}{2} \sqrt{ \Delta t^2 \Omega_j (4 - \Delta t^2 \Omega_j)}.
\]

To show \( |\lambda_{j_{1,2}}| = 1 \), we consider two cases:

\begin{enumerate}
    \item \textbf{If } \( \Delta t^2 \Omega_j = 4 \), then the square root vanishes and
    \[
    \lambda_{j_{1}} = \lambda_{j_2} = -1, \quad |\lambda_{j_{1,2}}| = 1.
    \]

    \item \textbf{If } \( \Delta t^2 \Omega_j < 4 \), then the eigenvalues are complex conjugates. Their squared magnitude is
    \begin{align*}
    |\lambda_{j_{1,2}}|^2 &= \left( \frac{2 - \Delta t^2 \Omega_j}{2} \right)^2 + \left( \frac{1}{2} \sqrt{ \Delta t^2 \Omega_j (4 - \Delta t^2 \Omega_j)} \right)^2 \\
    &= \frac{(2 - \Delta t^2 \Omega_j)^2 + \Delta t^2 \Omega_j (4 - \Delta t^2 \Omega_j)}{4} \\
    &= \frac{4 - 4 \Delta t^2 \Omega_j + \Delta t^4 \Omega_j^2 + 4 \Delta t^2 \Omega_j - \Delta t^4 \Omega_j^2}{4} = 1.
    \end{align*}
\end{enumerate}

Thus, in both cases, \( |\lambda_{j_{1,2}}| = 1 \), completing the proof.
\end{proof}

% {
% Propositions A.1, A.2, and A.3 are adapted with modifications from Propositions 3.1, 3.2, and E.1, respectively, from Rusch and Rus \cite{rusch2025oscillatory}.}

\subsection{Parallel scan in SHaRe-SSM}
% \citep{higuchi2024understanding} demonstrated that a gradient flow framework can uncover the local state dynamics of individual neurons. In particular, the resonator state $s_n = (u_n, v_n)$ can be expressed explicitly as a function of its previous state $s_{n-1}$. This recursive formulation enables efficient modeling of neuronal dynamics through parallelizable operations such as scan functions.

Parallel scans~\citep{kogge1973parallel, blelloch1990prefix} exploit associativity to reduce recurrent computation from $\mathcal{O}(N)$ to $\mathcal{O}(\log N)$. Originally developed for RNNs, they have recently been adapted to state-space models~\cite{s5}, enabling efficient architectures such as LRUs~\cite{lru} and Mamba~\cite{mamba}. In our setting, parallel scans accelerate linear updates, with spike functions applied afterwards. Following~\citet{rusch2025oscillatory}, we define an associative binary operation:  
\begin{equation}
    (a_1, a_2) \bullet (b_1, b_2) = (b_1 \cdot a_1,\; b_1 \cdot a_2 + b_2),
\end{equation}
where $\cdot$ denotes matrix-matrix or matrix-vector multiplication. Applying a parallel scan to the input sequence $\{(M,F_n)\}$ efficiently solves  
\begin{equation}
    s_n = Ms_{n-1} + F_n,
\end{equation}
with the second tuple element storing $x_n$. Efficiency is achieved by exploiting structured matrices (e.g., diagonal block $2 \times 2$ forms in $M_{IM}$ and $M_{IMEX}$), where each multiplication is linear in the hidden dimension. We use this formulation to implement IM and IMEX discretisations for HRF neurons. Algorithm~\ref{alg: SHaRe} summarises the SHaRe-SSM implementation.  

\subsection{Energy Computation}
We compute the energy for a LinOSS block and compare it to our SHaRe-SSM block. We can observe that our block doesn't perform any matrix multiplications and is well-suited for neuromorphic hardware. For event-based sensors, we can detach the encoder head and feed data directly to the model for real-time sequential processing. Spike rates from linear layers, SHaRe-SSM neuron and post weighted spike-mixing layers are given by $f^\theta, f^{\theta_C}, f^{\theta_D}$ for Sequence Length(L), State Size(P), Hidden Size(H), respectively.\\ 
The ratio of Energy consumed by LinOSS/SHaRe-SSM computed by: 
\[
\textstyle
\text{Ratio}=\frac{
    E_{MAC}\times N\left( 
        {2 L P  H} + 
        {(7+2) L  H^2} 
    \right)
}{
    E_{AC} \sum_{i=1}^{N} \left(  
    {\left(\sum_{j=1}^{i}f^{\theta}_j+f^{\theta_C}_i\right) L P  H}_{B,C} + 
        {f_i^{\theta_D}L H^2} 
    \right)
}
\]

We estimate the theoretical energy consumption of our model based on prior works~\citep{spikingssm, pspikessm, wu2025spikf}. Accordingly, we assume that MAC and AC operations are implemented on a 45nm hardware~\citep{horowitz20141}, where $E_{\text{MAC}} = 4.6\text{pJ}$ and $E_{\text{AC}} = 0.9\text{pJ}$. Notably, as discussed in ~\citeauthor{spikingssm}, the computational cost of multiplication of a floating-point weight by a binary activation number is represented as an addition-only operation. For ANNs, the theoretical energy consumption of a block $n$ is given by
$4.6\text{pJ} \times \text{FLOPs}(n)$. 
For SNNs, the energy consumption for $n$ is given by
$0.9\text{pJ} \times \text{SOPs}(n)$. 
Calculating theoretical energy consumption requires first calculating the synaptic operations, $\text{SOPs}(n) = f_r \times \text{FLOPs}(n)$, $f_r$ is the firing rate of the input spike train of the block/layer, $\text{FLOPs}(n)$ refers to the number of floating-point operations in layer $n$, equivalent to the number of multiply-and-accumulate (MAC) operations. SOPs denote the number of spike-based accumulate (AC) operations.

Since $\Omega$ is a diagonal matrix, it can be efficiently implemented using parallel scans with computations of order $\mathcal{O}(P \, \log(L))$ (Total computations are $\mathcal{O}(PL)$), which is negligible. Also, D is $\mathcal{O}(HL)$, which is much smaller than $\mathcal{O}(LPH)$, i.e., computation for B \& C matrices or even $\mathcal{O}(LH^2)$, which is for the GLU layer. \cite{rusch2025oscillatory} uses GeLU and GLU non-linearities, which we replace with a linear layer. And, as described in \citep{yu2023metaformer}, GeLU consumes 14 FLOPs per operation. Moreover, GLU has twice as many FLOPs as a linear layer.

\begin{table*}[ht]
\caption{Hyperparameters for the Best model for each dataset}
\label{tab: hyperparameters}
\centering
% \scalebox{0.5}{
\begin{tabular}{|l|l|c|c|c|c|c|c|}
\hline
Dataset & Method & LR & Hidden & State & Blocks & Time & Kernel \\
\hline
UCI-HAR     & IM    & 1e-3   & 128 & 256  & 2 & False &-\\
UCI-HAR     & IMEX    & 1e-3   & 128 & 256  & 2 & False &-\\
\hline
SHAR     & IM    & 1e-3   & 128 & 256  & 2 & False &-\\
SHAR     & IMEX    & 1e-3   & 128 & 256  & 2 & False &-\\
\hline
% Worms     & IM    & 1e-3   & 128 & 64  & 2 & False &-\\

Worms     & IM    & 1e-4   & 128 & 256  & 2 & False &-\\
Worms     & IMEX  & 1e-3   & 128 & 64  & 2 & False  &-\\
% Worms     & IM    & 1e-4   & 128 & 256  & 2 & False &-\\
\hline
% PPG       & IMEX    & 1e-3   & 128  & 256  & 4 & False & 16\\
PPG       & IM    & 1e-3   & 64  & 64  & 6 & False & 8\\
PPG       & IMEX    & 1e-3   & 128  & 256  & 6 & False & 16\\

\hline
\end{tabular}
\end{table*}
\end{document}